\documentclass{article}
\usepackage{iclr2027_conference,times}
\usepackage{amsmath,amssymb,amsthm}
\usepackage{graphicx}
\usepackage{booktabs}
\usepackage{array}
\usepackage{longtable}
\usepackage{placeins}
\usepackage{hyperref}
\usepackage{url}
\usepackage{xurl}
\usepackage{tabularx}  
\usepackage{makecell}  

\newcommand{\fullmark}{\ensuremath{\checkmark}}
\newcommand{\partialmark}{\ensuremath{\circ}}
\newcommand{\nomark}{\textemdash}

\usepackage{xcolor}
\title{MedImageOSWorld: Benchmarking GUI Agents for Medical Image Consoles}

\iclrfinalcopy
\author{\parbox{0.96\textwidth}{\normalfont\small\raggedright
\textbf{Ziyang Long}\textsuperscript{1,2,*}\quad
\textbf{Xinqi Li}\textsuperscript{3,4,*}\quad
\textbf{Lujing Xing}\textsuperscript{1,2,*}\\[3pt]
\textbf{Hsin-Jung Yang}\textsuperscript{1,\ensuremath{\dagger}}\\[6pt]
\textsuperscript{1}Biomedical Imaging Research Institute, Cedars-Sinai Medical Center, Los Angeles, CA, USA\\[2pt]
\textsuperscript{2}University of California, Los Angeles, Los Angeles, CA, USA\\[2pt]
\textsuperscript{3}Technical University of Munich\\[2pt]
\textsuperscript{4}Berlin Ultrahigh Field Facility, Max Delbr\"uck Center for Molecular Medicine in the Helmholtz Association, Berlin, Germany\\[4pt]
\textsuperscript{*}Equal contribution (Co-first Authors).\quad
\textsuperscript{\ensuremath{\dagger}}Corresponding author: \href{mailto:hsin-jung.yang@csmc.edu}{\texttt{hsin-jung.yang@csmc.edu}}.
}}

\begin{document}
\maketitle
\lhead{Preprint}

\begin{abstract}
  Graphical consoles offer a practical interface for medical acquisition
  assistance, allowing agents to work through the controls and visual feedback
  used by human operators. Reliable assistance requires linking on-screen
  anatomy to acquisition decisions that determine what image evidence becomes
  available next. We introduce MedImageOSWorld, a benchmark for evaluating this
  capability in simulated CT, MR, and ultrasound consoles. Using screenshots and mouse-and-keyboard actions, agents configure protocols, plan acquisitions, inspect the resulting images, and make corrective adjustments across seven capability levels, from console operation to feedback-driven control.
  Evaluation combines task-specific workflow checks with hidden anatomical
  ground truth to assess procedural completion and acquisition outcomes
  separately. A common evaluation protocol specifies episode conditions and
  interaction budgets, while recorded trajectories support analysis of how
  agents observe, act, and respond to acquisition feedback. Across eleven
  open-weight agents, success rates range from 3.0 to 25.0 on a 0--100 scale
  while workflow-progress rates reach 21.5--74.5: agents complete much of the
  console workflow but rarely acquire the intended anatomy. Success collapses
  between perception and planning, from 79--90\% at the lowest three levels for
  the best agent to at most 12\% for millimetre-level planning and 0\% for
  closed-loop control among open-weight agents. Two proprietary agents reach success rates of about 34
  and exceed the best open-weight agent mainly in acquisition quality (61
  versus 42). MedImageOSWorld provides a controlled setting for studying
  whether general-purpose GUI agents can translate visual observations into
  effective medical acquisition decisions. 
\end{abstract}

\section{Introduction}
\label{sec:introduction}

Medical image acquisition depends on operators translating an examination
request into console actions. They configure protocols, plan anatomical
coverage or imaging planes, and assess the resulting images. The console is the interface where an operator turns an examination goal
into acquired image evidence. An inadequate
acquisition may require adjustment, repetition, or escalation. 

The graphical console is a natural interface for acquisition assistance: it
brings protocol settings, spatial planning, and image assessment into the
operator's existing workflow. Agents that use screenshots and standard input
actions can interact through this interface without requiring a dedicated
scanner-control API or direct access to internal acquisition state. This offers
a route to automation across heterogeneous console software while keeping
actions and visible outcomes available for operator review. It also creates a
demanding control problem: the agent must connect what is shown on screen to
acquisition decisions whose consequences appear in subsequent images.

Recent advances in general-purpose GUI agents make this interaction setting
increasingly relevant. Computer-use benchmarks evaluate multimodal agents on
stateful workflows over desktop applications and web services
\citep{yuan2026osworld2}. Medical acquisition poses a further question: can
such agents combine visual understanding, spatial reasoning, and
feedback-driven control to acquire the intended anatomy without human
correction during an episode? We ask not only whether an agent can operate the
machine, but whether its actions produce an acquisition that meets the intended
goal.

The distinction arises because acquisition actions determine what evidence
becomes available for subsequent decisions. We call this property
\emph{action-conditioned acquisition}: an agent action alters the acquisition
process and thereby changes the medical-image evidence produced at subsequent
steps. The agent must identify the anatomy on screen, relate image coordinates
to acquisition geometry, and use the resulting evidence to decide whether to
proceed or adjust. Changes to geometry, timing, or probe pose affect acquisition;
display-only adjustments do not. An agent can complete the prescribed workflow
and still acquire the wrong anatomy. Workflow completion alone does not
establish that the intended anatomy was acquired.

Existing environments address parts of this evaluation problem. Clinical
benchmarks cover electronic health records, simulated encounters, and
administrative workflows
\citep{jiang2025medagent,schmidgall2024agentclinic,bedi2026healthadmin}, while
MedCUA-Bench evaluates clinical GUI interaction with safety-critical errors
\citep{yu2026medcua}. These settings primarily concern records, software state,
or predefined observations rather than imaging acquisition. ABRA, RadABench,
and MedSPOT study post-acquisition viewing, radiology tool use, and interface
grounding \citep{maksudov2026abra,zheng2024radabench,shakeel2026medspot}; they
do not let the agent control the acquisition that generates the underlying
study. Autonomous MRI and MaRGE/MaRCoS address acquisition and scanner control
\citep{ravi2020autonomous,algarin2024marge}, but are MRI-specific systems rather
than standardized benchmarks for general-purpose GUI agents. The remaining
need is an evaluation setting that connects executable console workflows to
action-conditioned acquisition and independent anatomical ground truth.

\begin{figure}[t]
  \centering
  \includegraphics[width=\textwidth]{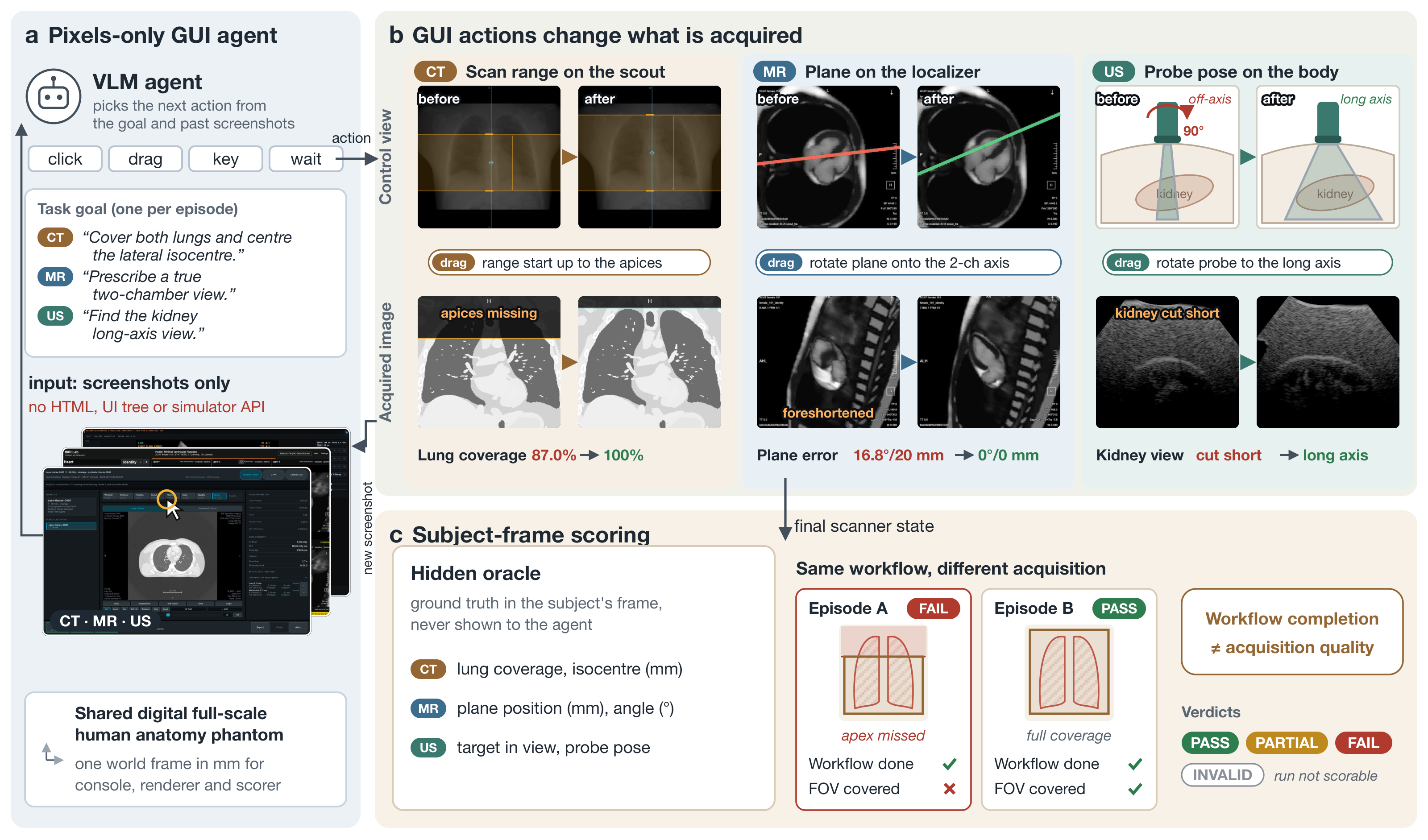}
  \caption{\textbf{MedImageOSWorld evaluation loop.} (a) A GUI agent receives console
  screenshots and acts through mouse and keyboard inputs, without interface or
  simulator state. (b) Changes to CT scan range, MR plane, or ultrasound probe
  pose alter the image evidence acquired next. (c) A hidden oracle evaluates
  the final acquisition in the subject's coordinate frame, separating workflow
  completion from acquisition quality. The displayed measurements illustrate
  the scoring interface and are not experimental results.}
  \label{fig:overview}
\end{figure}

To study this question, an environment must preserve the dependencies between
planning, acquisition, and assessment while retaining ground truth for the
acquired result. We introduce \textbf{MedImageOSWorld}, a benchmark with three
executable consoles: computed tomography (CT), magnetic resonance (MR), and ultrasound. All three derive observations from
XCAT-based digital human anatomy phantoms \citep{segars2010xcat}. The environment couples
an order, a screening record, persistent patient and device state, a protocol
library, and an action-conditioned acquisition backend. Agents select and
configure protocols, plan anatomy-dependent slices, volumes, or views, acquire
images, and assess their adequacy. They can re-plan, re-acquire, repeat a view,
or escalate. Figure~\ref{fig:overview} separates the three stages that define
an episode: visible console interaction, a hidden physical-state transition,
and outcome evaluation against anatomical ground truth.

MedImageOSWorld provides a setting for studying how general-purpose agents
combine visual understanding, spatial reasoning, and feedback-driven control
in tasks with simulated physical consequences. A shared capability ladder
organizes these requirements and helps locate failure points. Evaluation
separates workflow completion from acquisition quality using hidden ground
truth in the digital subject's coordinate frame. This makes the target of
evaluation the connection between interface actions and measurable acquisition
outcomes, rather than procedural completion alone. Our contributions are:
\begin{itemize}
  \item \textbf{Action-conditioned acquisition as an evaluation setting.} We
  formalize medical image acquisition as a GUI-agent task in which actions
  change the physical acquisition state and therefore the image evidence
  observed next. This setting separates two questions that existing
  benchmarks conflate: whether the agent completed the console workflow, and
  whether the resulting acquisition captured the intended anatomy.
  \item \textbf{MedImageOSWorld, a benchmark of executable scanner consoles.}
  Three simulated consoles (CT, MR, ultrasound) expose protocol selection,
  geometric planning, acquisition, review, and corrective actions through
  rendered screenshots and a constrained mouse-and-keyboard interface, with no
  access to interface or simulator state. 
  \item \textbf{Ground-truth scoring in the subject's coordinate frame.} Each
  committed acquisition is linked to the digital subject's coordinate frame,
  so a hidden oracle can grade geometry and
  anatomical coverage in millimetres, degrees, and coverage fractions
  without human assessment. Success, workflow progress, and acquisition quality
  are reported as three separate 0--100 scores, with environment faults kept
  apart as \texttt{INVALID}.
  \item \textbf{An empirical study of current GUI agents.} We evaluate
  11 open-weight models and two API-served proprietary models on 31,197 valid
  episodes included in scoring (30,513 for open-weight agents and 684 for
  proprietary agents) under a frozen interaction and scoring protocol. The best
  open-weight agent, Qwen3.8-27B, reaches a success rate of 25.0 but a
  workflow-progress rate of 74.5; no open-weight agent exceeds 12\% success on
  millimetre-level planning, and CT remains nearly unsolved, with success at or
  below 1.5 for every open-weight agent. The simulator and data will be  available on
\href{https://huggingface.co/datasets/neutral-123/MedImageOSWorld}{Hugging Face.}
\end{itemize}

\section{Related Work}
\label{sec:related-work}
\textbf{General computer-use benchmarks.}
Table~\ref{tab:environment-comparison} contrasts representative clinical,
radiology, and acquisition environments across the interaction, acquisition,
and evaluation properties that define this task setting. OSWorld evaluates
multimodal agents by executed outcomes in real computer environments, and
OSWorld 2.0 extends this line to longer-horizon tasks
\citep{xie2024osworld,yuan2026osworld2}. AndroidWorld varies task parameters in
a dynamic mobile environment, while WebArena evaluates functional correctness
on realistic web tasks \citep{rawles2024androidworld,zhou2023webarena}. Like
these benchmarks, we score executed outcomes rather than action traces and use
task-specific parameter variation; MedImageOSWorld additionally introduces a
physical acquisition state that determines the next observation.

\textbf{Clinical and radiology agents.}
MedAgentBench and AgentClinic study EHR use and simulated encounters
\citep{jiang2025medagent,schmidgall2024agentclinic}; the health administration
benchmark and MedCUA-Bench cover administration and clinical GUI interaction
\citep{bedi2026healthadmin,yu2026medcua}. RadABench, MedSPOT, and ABRA study radiology
tool use, interface grounding, and DICOM viewing
\citep{zheng2024radabench,shakeel2026medspot,maksudov2026abra}. In these
settings, the relevant records or imaging study are fixed before the episode;
the agent does not control how the evidence is acquired. MedImageOSWorld adds
this acquisition loop before review or interpretation.

\textbf{Autonomous acquisition and digital subjects.}
Autonomous MRI supports automated MRI protocolling, planning, acquisition, and
reconstruction, while MaRGE/MaRCoS supports research MRI control
\citep{ravi2020autonomous,algarin2024marge}. These are MRI-specific acquisition
systems rather than benchmarks for general-purpose GUI agents. MedImageOSWorld
links console actions and committed acquisitions to the subject's coordinate
frame so that geometry and anatomical coverage can be scored automatically.

\FloatBarrier
\section{The MedImageOSWorld Environment}
\label{sec:method}
We model each episode as a partially observed interaction between an agent and a simulated console. The console state has two components. The \emph{physical acquisition state} determines what the scanner would acquire, such as the CT slab and isocentre, the MR imaging plane, or the ultrasound probe pose and tissue contact. The \emph{interface and workflow state} records the worklist, sequence queue, workflow stage, and display settings. At each step, the agent issues a single GUI action; the simulator then updates the console state using task-relevant physical abstractions. Display-only actions, such as adjusting window width and level, modify the interface state but leave the physical acquisition state unchanged. Section~\ref{sec:ground-truth} describes the separate evaluation of workflow completion and acquisition quality. Figure~\ref{fig:overview} illustrates this loop.

\begin{table}[t]
\centering
\footnotesize
\setlength{\tabcolsep}{3pt}
\renewcommand{\arraystretch}{1.05}
\renewcommand{\cellalign}{cc}
\caption{Comparison of environments related to GUI agents and medical image
acquisition.}
\label{tab:environment-comparison}
\begin{tabularx}{\textwidth}{@{}l>{\raggedright\arraybackslash}X*{4}{c}@{}}
\toprule
\textbf{Work} & \textbf{Setting} &
\textbf{\makecell{Pixels-only\\obs.}} &
\textbf{\makecell{Plans\\acq.}} &
\textbf{\makecell{Action-dep.\\image}} &
\textbf{\makecell{Subject-frame\\GT}} \\
\midrule
\makecell[l]{OSWorld 2.0\\\citep{yuan2026osworld2}}       & Desktop and web computer use     & \fullmark    & \nomark   & \nomark   & \nomark \\
\makecell[l]{MedAgentBench\\\citep{jiang2025medagent}}    & FHIR-based EHR tasks             & \nomark      & \nomark   & \nomark   & \nomark \\
\makecell[l]{RadABench\\\citep{zheng2024radabench}}       & Prompt-driven radiology tools    & \nomark      & \nomark   & \nomark   & \nomark \\
\makecell[l]{ABRA\\\citep{maksudov2026abra}}              & DICOM viewer and reporting tools & \partialmark & \nomark   & \nomark   & \nomark \\
\makecell[l]{MaRGE/MaRCoS\\\citep{algarin2024marge}}      & Research MRI scanner console     & \nomark      & \fullmark & \fullmark & \nomark \\
\makecell[l]{Autonomous MRI\\\citep{ravi2020autonomous}}  & Automated brain MRI workflow     & \nomark      & \fullmark & \fullmark & \nomark \\
\midrule
\textbf{MedImageOSWorld (ours)} & \textbf{Simulated CT, MR, and US consoles} &
\textbf{\fullmark} & \textbf{\fullmark} & \textbf{\fullmark} & \textbf{\fullmark} \\
\bottomrule
\end{tabularx}

\vspace{2pt}
{\scriptsize\raggedright\textit{Notes.} ``Action-dep.\ image'' means that the
agent's choices affect subsequently acquired image evidence, beyond changing
the GUI display or navigating a fixed study.
\fullmark: full support; \partialmark: partial or indirect support;
\nomark: absent or not applicable. ABRA provides image pixels but controls the
viewer through function-calling tools, so its observation is not pixels-only.\par}
\end{table}

\subsection{Observation, action, and time}
The primary \textit{observation} is a PNG screenshot of the console rendered at a fixed viewport. The \textit{action} space consists of constrained mouse events and an allowlisted set of keyboard events, so the agent interacts through the same visible console and standard input modalities available to a human operator. The agent receives no DOM, accessibility tree, simulator or interface state, script-execution primitive, hidden UI-quality text, or direct scanner-control API; task geometry and the consequences of each action are therefore available only through the rendered pixels.
Before execution, a declared, version-fixed, deterministic codec may normalize model output into a legal action. For every step, an action receipt records the raw model output, any normalization or repair applied, whether the action was executed or rejected, and any additional feedback returned to the agent. Rejected actions consume steps as specified by the frozen interaction profile, and any model-specific repair logic or additional prompting is disclosed as part of the agent configuration. Simulated \textit{time} advances only through executed actions, including explicit wait actions, rather than during model inference.
\subsection{Three consoles}
All three consoles share a common structure (Figure~\ref{fig:overview}b):
planning controls modify the physical acquisition state, acquisition commits
it, and the acquired images provide evidence for assessment and correction.

\textbf{Computed tomography (CT).} After worklist and protocol selection, the agent plans on a scout image by
setting the scan slab, isocentre, monitoring region of interest (ROI), and
acquisition timing. The acquired stack is presented separately from the
prescription, so undercoverage can be detected and corrected. Task families
cover thoracic, renal, hepatic, and cardiac acquisition, including contrast
timing, repeat acquisition of clipped anatomy, and coverage under a dose-index
constraint.

\textbf{Magnetic resonance (MR).}
The agent prescribes imaging planes and volumes on localizer images by setting
plane position, orientation, field of view, and slab or slice coverage; each
acquisition reslices the digital subject at the committed geometry. Within the
sequence queue, acquired images guide the refinement or repetition of
subsequent prescriptions for cardiac cine planes, cardiac T1/T2 mapping, and
bilateral renal coverage.

\textbf{Ultrasound (US).}
The live renal image is rendered from the current probe position, orientation,
and tissue contact, and from the depth, frequency, and focus settings. Tasks
require searching for a kidney view, adjusting acquisition parameters,
distinguishing contact loss from inadequate gain, and freezing and saving a
selected frame.

\subsection{Digital subjects and ground truth}
\label{sec:ground-truth}
Benchmark episodes use virtual subjects in simulated consoles and involve no
scanning of humans or animals.
The digital anatomy phantoms provide registered anatomical structures in a
shared subject coordinate frame, in which every planning control and committed
acquisition is expressed. CT slabs and isocentres, MR planes and volumes, and
ultrasound probe planes can therefore be compared directly with hidden
anatomical masks and landmarks, enabling automatic measurement of endpoint and
centre error, plane position and orientation, field of view, and anatomical
coverage, independent of screen coordinates or click sequences.

Workflow checks use the interface state and event log to verify that required
stages and actions occurred, whereas the hidden oracle evaluates the anatomy
actually acquired; a workflow can thus complete without meeting the
acquisition-quality criteria. Rendering assumptions, coordinate conventions,
and modality-specific scope are documented in the supplementary section \textit{Simulator Rendering and Physical-Model Scope}.

\FloatBarrier
\section{Tasks and Capability Ladder}
\label{sec:ladder}

\begin{figure*}[t]
  \centering
  \includegraphics[width=\textwidth]{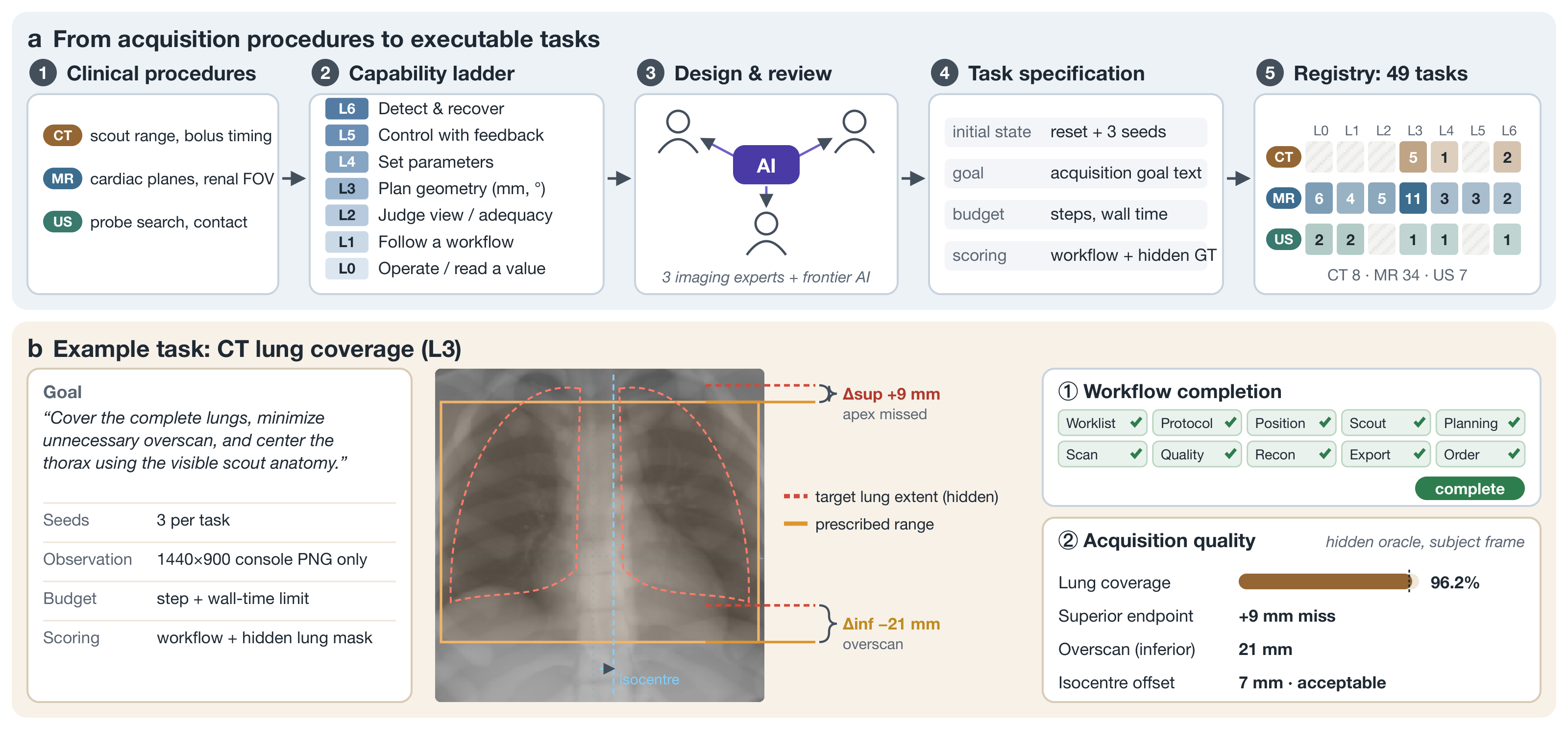}
  \caption{\textbf{From acquisition procedures to executable benchmark tasks.}
  (a) The shared L0--L6 capability ladder is defined first; three medical
  imaging experts, assisted by a frontier AI model, propose, cross-review, and
  finalize each task. Each task specifies an initial state, an acquisition
  goal, an interaction budget, and scoring conditions. (b) A CT lung-coverage
  task links scout-based planning to the acquired volume, with workflow
  completion and anatomical coverage scored separately. Numbers in (b) are
  illustrative.}
  \label{fig:task-construction}
\end{figure*}

\paragraph{Task sourcing.}
We first defined the capability ladder and then designed tasks against it
(Figure~\ref{fig:task-construction}a). Three medical imaging experts, assisted
by a frontier AI model, proposed, cross-reviewed, and finalized each task.
The benchmark contains 49 task definitions: 8 for CT, 34 for MR, and 7 for
ultrasound. They cover CT coverage planning,
dose constraints, and undercoverage recovery; MR console operation, acquisition
planning, mapping, and coverage assessment; and ultrasound kidney operation,
measurement, search, optimization, and fault recovery. The tasks evaluate
acquisition operations, not diagnostic accuracy.
The supplementary section ``Task Catalog and Reproducibility Details'' lists
every task with a short description and its capability level; execution coverage
is reported separately from task membership.

\paragraph{Capability ladder.}
A task's level is the capability it targets, not the interaction mechanism it
uses (Figure~\ref{fig:task-construction}a). L0--L1 concern operating the console,
L2 concerns interpreting the current observation, and L3--L6 add spatial
quantification, physical constraints, temporal closure, and recovery. A
composite task is counted once, at its primary level; for example,
\path{ct_ui_smoke_routine} completes the full routine workflow but is placed at
L3 because it is gated on lung geometry. The ladder helps locate where
performance breaks down, but it is not a psychometrically validated scale, and
a single failure does not identify a unique cause: a planning failure may
originate in an earlier perception error.

\paragraph{Task specification.} Each task is defined by its goal text, modality, capability level, reset
distribution, observation and action profile, step and wall-time budgets,
termination conditions, ground-truth scorer, and task and scorer versions. A
task may define multiple conditions, and each condition targets 30 digital
human subjects with three seeds per subject. Proprietary agents are evaluated
on the first five assets of each task in the frozen order. The episode
manifest distinguishes task definitions, conditions, planned slots, attempts,
and scored episodes, and records the subject, seed, bundle, and reset
information associated with each.

\section{Evaluation Protocol}
\label{sec:scoring}

\paragraph{Verdicts.}
The task scorer assigns \texttt{PASS}, \texttt{PARTIAL}, \texttt{FAIL}, or an
undetermined outcome. Environment, receipt, or scorer faults are
\texttt{INVALID}. Slots left unrun after a confirmed model-side early stop are
counted as failures in SR. \texttt{INVALID} episodes and slots lost to
infrastructure faults are excluded from metric denominators and reported
separately.

\paragraph{Workflow and acquisition are scored separately.}
Workflow checks cover required and forbidden events and the terminal state.
Acquisition quality is measured against hidden ground truth in the digital
subject's frame: geometry in millimetres and degrees, and coverage as the
fraction of the target structure inside the acquired volume. Coverage is
computed from the acquired slab, never inferred from the prescribed range or a
completion button (Figure~\ref{fig:task-construction}b). Thresholds are task-
and scorer-version specific; for the MR \texttt{L4\_2ch} task, the current
relaxed engineering profile gates centre, normal, field of view, and landmark
errors at 8~mm, 5~degrees, 10~mm, and 5~mm. Continuous components remain
visible alongside every verdict.

\paragraph{Protocol freeze and aggregation.}
A valid comparison freezes the task version, subject or bundle, reset,
observation and action profile, budgets, scorer and thresholds, runtime, and
agent configuration. Task-side manifests are shared across compared agents;
agent-side prompts, codecs, deterministic normalizations, repair policies, and
extra feedback may differ and are disclosed. Step accounting follows the
frozen profile for rejected outputs, repairs, and corrective feedback;
transport-level retries are logged separately. Results are laid out as model
$\times$ modality $\times$ level
$\times$ observation condition, with empty modality--level cells shown as N/A.
We report three 0--100 metrics. Success rate (SR) is the percentage of
registered slots that satisfy the task's complete success criterion. Progress
rate (PR) is the mean percentage of registered workflow progress completed per
slot. Acquisition quality (AQ) is the mean registered, normalized quality of
the acquired result, computed from task-specific hidden-ground-truth measures.
The legacy \texttt{PARTIAL} verdict provides no fractional credit to any main
metric. Task-specific geometry, coverage, gate values, exhausted-budget rates,
and the counts and reasons for \texttt{INVALID} episodes are reported alongside
SR, PR, and AQ.

Conditions are weighted equally within each task; the clean and defect
conditions of MR renal adequacy each contribute half of its task score.
Condition-specific results include answer frequencies, no-answer rates, and
actual denominators after the exclusions defined above. Intervals resample
subjects with their repeated runs grouped.
The official scorer identity is the tuple of scorer code digest, oracle digest,
and threshold version; changing any element creates a new scorer version.

\paragraph{Overall score.}
Let $s_{ij}$ be a model's SR proportion for task $j$ at level $i$, after
condition aggregation and frozen denominator rules. The L0--L6 aggregate uses
$w_i=i+1$:
\begin{equation}
  \mathrm{Score}=100\frac{\sum_{i=0}^{6}(i+1)\sum_{j=1}^{n_i}s_{ij}}
                           {\sum_{i=0}^{6}(i+1)n_i},\qquad 0\leq s_{ij}\leq1.
  \label{eq:overall-score}
\end{equation}
Here $n_i$ counts task definitions; the current denominator is 181. Higher
levels receive greater pre-selected weight, not empirically validated
psychometric difficulty. Modality and level scores are also reported. The
pooled SR score is not modality-equal. SR, PR, and AQ are reported side by side
and never merged.

\section{Experiments}
\label{sec:experiments}

\subsection{Setup}

We ask four questions. Which registered requirements can current GUI agents
satisfy across modalities and capability levels? How often does workflow
completion coincide with acquisition quality? At which observation or
corrective gate do non-passing episodes stop making progress? How sensitive are
scores to the agent harness when the task-side protocol is held fixed?

We evaluate eleven open-weight GUI and vision-language agents:
Qwen3.8-27B, UI-Mate-27B, GUI-Owl-1.5-32B-Instruct,
GUI-Owl-1.5-32B-Thinking, Qwen3-VL-30B-A3B-Instruct,
Qwen3-VL-30B-A3B-Thinking, UI-TARS-1.5-7B,
Qwen2.5-VL-72B-Instruct, OpenCUA-32B, Gemma-4-31B, and Muse-Glimmer-30B.
All are served locally with greedy decoding (temperature 0). We additionally
evaluate OpenAI GPT-6 Sol (\texttt{gpt-6-sol}) and Anthropic Claude Opus 5.5
(\texttt{claude-opus-5-5}) through their provider APIs. 

The target design assigns each task condition 30 subjects $\times$ 3 seeds.
With clean and defect conditions, MR renal adequacy therefore has 180 planned
slots per model. The frozen manifest records actual attempts, outcomes,
exclusions, and unrun slots, together with the agent configuration for each
condition.

Hardware allocation and proprietary-agent harness versions are summarized in
the supplementary material. Full prompts, action-codec specifications, and
run configurations will be released through our
\href{https://huggingface.co/datasets/neutral-123/MedImageOSWorld}{Hugging Face repository}.

\subsection{Main results}

\begin{figure*}[t]
  \centering
  \includegraphics[width=\textwidth]{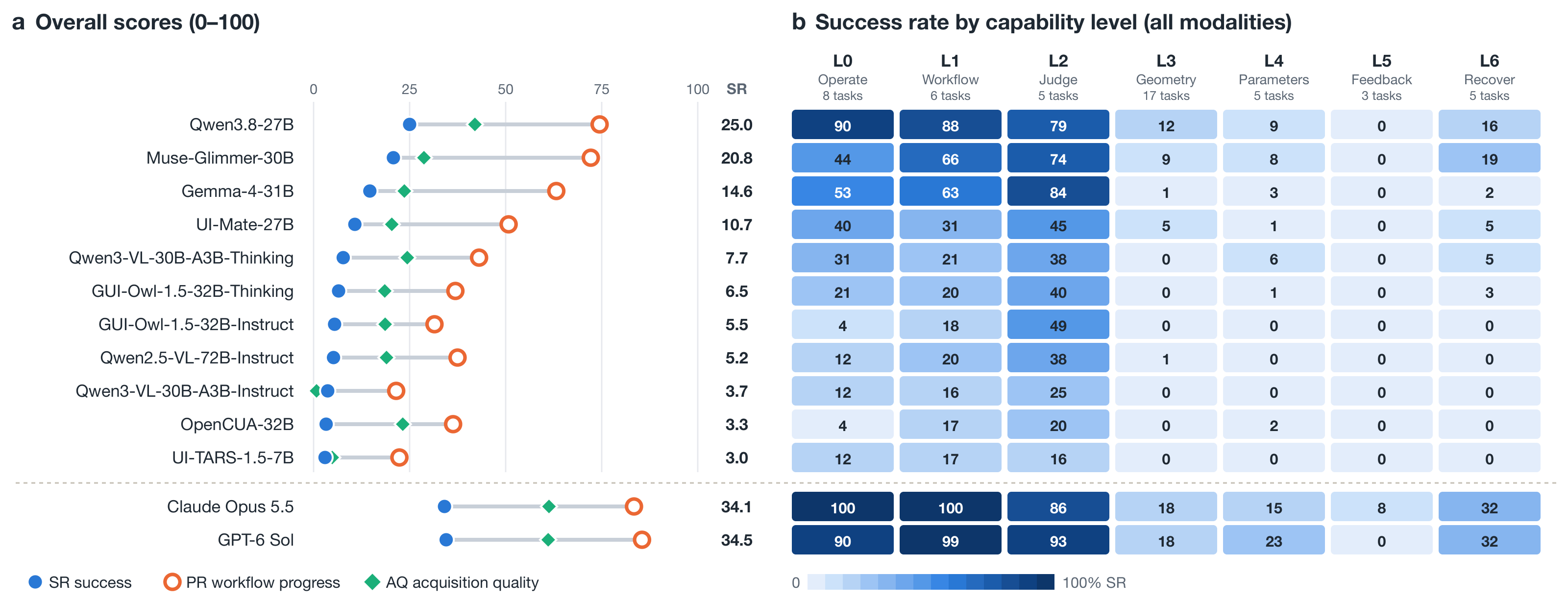}
  \caption{\textbf{Performance across tasks and capability levels.} (a) Overall
  success rate (SR; filled blue circles) is the primary metric: each scored
  episode receives 1 for \texttt{PASS} and 0 otherwise. Workflow progress (PR;
  open orange circles) measures completed workflow stages on a five-point scale
  (0, 0.25, 0.5, 0.75, or 1), without assessing acquisition quality.
  Acquisition quality (AQ; green diamonds) is a continuous score from 0 to 1
  for the 28 eligible geometry and parameter tasks. All three measures are
  shown as percentages; PR and AQ are auxiliary measures and are not combined
  with SR. Overall SR uses capability-level weights $w_i=i+1$ across 49
  tasks (weighted denominator 181). (b) Mean task-level SR within each
  capability level, pooled across CT, MR, and ultrasound; column headings show
  task counts. The divider separates locally served open-weight models from
  API-served proprietary models. API results use the first five assets per
  task.}
  \label{fig:capability-heatmap}
\end{figure*}

Figure~\ref{fig:capability-heatmap} summarizes performance across tasks and
capability levels. Panel (a) reports each model's overall SR, PR, and AQ, with
SR as the primary metric and PR and AQ shown as separate auxiliary measures.
Panel (b) reports mean task-level SR within each capability level, pooled across
CT, MR, and ultrasound. Overall SR uses Equation~\ref{eq:overall-score} across
49 tasks with a weighted denominator of 181.

Among open-weight agents, Qwen3.8-27B leads with an SR of 25.0, followed by
Muse-Glimmer-30B (20.8), Gemma-4-31B (14.6), and UI-Mate-27B (10.7); the
remaining seven score between 3.0 and 7.7. Success drops sharply from L0--L2 to L3:
the best open-weight agent reaches 79--90\% at L0--L2 but at most 12\% at L3, and
every open-weight agent scores 0\% at L5. The proprietary agents reach SRs of
34.5 (GPT-6 Sol) and 34.1 (Claude Opus 5.5); their workflow progress is only
9--11 points above Qwen3.8-27B (85.5 and 83.4 versus 74.5), but their
acquisition quality is about 19 points higher (61.1 and 61.3 versus 42.0), and
CT remains difficult even for them (SR 10.3 and 1.7).

The CT console illustrates the gap between workflow progress and acquisition
quality. Qwen3.8-27B and UI-Mate-27B achieved median target coverage of
0.96 and 0.78, respectively, with median lateral centring errors of
6.7\,mm and 14.3\,mm.
Scaling the task tolerances from $0.5\times$ to $2\times$ changes the CT pass
rate. In the CT recovery tasks, every agent accepted the
repeat prompt and closed the study without correcting the clipped range (0
causal range corrections in the 360 Qwen3.8-27B and UI-Mate-27B episodes).
 
Coverage alone is also insufficient. On renal MR L6, Claude Opus 5.5 reached
full bilateral coverage in all 12 translation-defect cells but passed none.
In each case it restored coverage through a mechanism other than the
responsible setting, so the causal-change gate failed.

\subsection{Where agents fail}

We decompose each scored trajectory into observation gates and corrective
gates. A review gate records required viewing actions, such as opening an
image and visiting specified slices or windows; it does not score
interpretation. In renal tasks, \texttt{causal\_change} records a change to
the responsible acquisition parameter, and an independent coverage gate
measures the resulting anatomy. Other corrective gates cover geometry,
parameters, and terminal state.
 
\paragraph{Agents observe but do not correct.}
Observation gates are almost never the blocker. In renal MR tasks, agents
review the initial series or scouts in 100\% of episodes, and Claude Opus 5.5
satisfies every review gate in all 150 renal cells. Corrective gates are
rarely met. Across the eleven open-weight models (about 990 valid episodes
per renal task), the results are as follows:
\begin{itemize}
  \item In numeric coverage correction, agents change the responsible
  parameter in 14\% of episodes, and only 57\% of those changes restore
  bilateral coverage.
  \item In coverage recovery, the corresponding rates are 7\% and 36\%.
  \item Replacement acquisitions are completed in 25--50\% of episodes, but
  bilateral coverage is achieved in only 6--32\%.
\end{itemize}
On the CT console, 1,032 of 3,084 valid episodes
(33\%) never committed an acquisition. Among early-stopped arms, the
share of episodes reaching a committed acquisition ranged from 15\%
(OpenCUA-32B) to 49\% (Muse-Glimmer-30B). Gemma-4-31B and Muse-Glimmer-30B
exported a study in 42\% and 47\% of episodes and still passed none.
 
\paragraph{Budget exhaustion and looping.}
Overall, 64.6\% of scored open-weight episodes end with an exhausted action
budget; these account for 75.7\% of non-\texttt{PASS} episodes. By modality,
the share of episodes ending this way is 70.9\% in MR, 38.2\% in CT, and
29.0\% in ultrasound. On renal MR, exhaustion rises from 50\% at L2 to 84\%,
80\%, and 75\% at L3, L4, and L6. On CT, four agents repeated one identical
click for at least 200 of 400 steps in most episodes: Qwen2.5-VL-72B
(168/168), UI-TARS-1.5-7B (249/264), Qwen3-VL-30B-A3B-Instruct (193/216), and
OpenCUA-32B (125/228). These four agents exported in at most 2\% of episodes.
On ultrasound, the three zero-pass agents exhausted the 64-event budget in
89--100\% of episodes, compared with 1\% for Qwen3.8-27B. Whether
larger budgets improve success requires a separate budget-controlled comparison.
 
\paragraph{Primary failure reasons.}
Each non-\texttt{PASS} episode is assigned exactly one primary reason,
determined from its scorer gates, termination, rejected-action share (a
malformed-output loop has at least 50\% rejected actions), and wall time; no
case was adjudicated by hand. In the reference-harness condition of the renal
L6 recovery task (five agents, ten subjects, one seed, 50 episodes), the
counts are: exhausting the budget without a correction: 16 (32\%); declaring done without a correction: 12 (24\%); a malformed-output loop: 9 (18\%); reacquiring without changing the responsible setting: 4 (8\%); abstaining on a correctable case: 2; never declaring done on a clean case: 2; changing the responsible setting with coverage still incomplete: 1.
The remaining 4 episodes passed. Claude Opus 5.5 adds a further pattern,
over-correction: it altered a correct acquisition in all 10 clean L4 and L6
cells and abstained on all 30 clean L2 cells. Agents frequently reached the required review views without making an
effective correction.
 
The supplementary material illustrates how CT and MR observations and actions
can be aligned with workflow and acquisition-quality gates using schematic
trajectories; these are not recorded experimental episodes.

\section{Discussion}
\label{sec:discussion}

MedImageOSWorld provides a controlled setting for studying how general-purpose
multimodal agents apply their capabilities to specialized acquisition tasks.
The interaction primitives remain screenshots, clicks, drags, and keyboard
input, while successful actions depend on anatomy, acquisition geometry, and
the current subject. An agent must combine interface operation with domain
knowledge and visual reasoning to turn an acquisition goal into a working
prescription. The gap between workflow progress and task success makes this
combination a useful target for evaluating the transfer of general-purpose
agents to specialized visual workflows.

The results show where this combination currently breaks down. Success drops
sharply at geometric planning, where agents must translate two-dimensional
views into a position, orientation, and extent in the subject's coordinate
frame. Their acquisition choices then determine the images available for the
next decision. In renal and CT recovery tasks, agents frequently reach the
review stage without making an effective adjustment; some also alter an
adequate acquisition in clean cases. Together, these findings point to
difficulties in connecting visual evidence to the choice and verification of
a corrective action. The logs do not isolate whether a failure begins with
anatomical interpretation, coordinate mapping, control selection, or
execution. Continuous geometry and coverage measurements help locate the
resulting error, while task-specific gates record whether the required
recovery procedure was followed.

This setting also allows future evaluations to ask whether general model
upgrades improve acquisition outcomes without additional domain
demonstrations, and how much a small number of expert demonstrations can
help. Comparing these conditions on held-out subjects and task variations
would test how readily agents adapt their existing capabilities to acquisition
workflows. Improvements should be assessed through acquisition quality and
task success alongside workflow progress, especially on planning and recovery
tasks. Continuous outcome measures can detect useful changes before an
acquisition crosses a binary success threshold. Such comparisons need fixed
task and scorer versions, subjects, reset conditions, and budgets, with model
revisions and harness changes recorded separately.

Interpreting these comparisons requires accounting for both training exposure
and evaluation conditions. The evaluated models' prior exposure to related
interfaces or procedures is not established, so the present results do not
isolate generalization to unseen training domains. Proprietary agents also
use fewer subjects and provider-specific harnesses, which limits attribution
of their advantage to the underlying models. The simulated consoles measure
selected anatomical and parameter objectives; transfer to scanner hardware,
diagnostic image quality, and the benefit of human oversight require further
evaluation. Within these limits, the benchmark makes it possible to examine
whether gains in general-purpose agents translate into more effective
acquisition decisions.

\section{Conclusion}
\label{sec:conclusion}
MedImageOSWorld evaluates general-purpose GUI agents on 49 medical acquisition
tasks across simulated CT, MR, and ultrasound consoles. Current agents often
advance through the workflow while failing to achieve the required geometry
or make effective corrections. By linking interface actions to acquired images
and hidden anatomical ground truth, the benchmark measures how visual
understanding is translated into acquisition outcomes. It provides a
controlled setting for studying whether improvements in multimodal agents
carry over to specialized tasks that require spatial planning and
feedback-driven action.

\section*{Code and Data Availability}

The simulator, data, and reproducibility materials will be released through our
\href{https://huggingface.co/datasets/neutral-123/MedImageOSWorld}{Hugging Face repository}.
The release will include prompts, action-codec specifications, run configurations
and manifests, simulator and scorer revisions, episode logs, and analysis
scripts, with access instructions for separately licensed assets.

\section*{Acknowledgments}

Generative AI tools, including Claude Code and Codex CLI, assisted with
manuscript preparation, literature searches, software implementation,
experiment orchestration, and analysis of experimental reports. SR, PR, and AQ
were computed from episode logs by deterministic scoring scripts; no AI model
directly assigned scores. The authors reviewed and tested the generated code,
checked reported results against the experimental data, and take responsibility
for the final manuscript and artifacts.

\bibliographystyle{iclr2027_conference}
\bibliography{references}

\end{document}